\documentclass{article}

\usepackage[english]{babel}

\usepackage[letterpaper,top=2cm,bottom=2cm,left=3cm,right=3cm,marginparwidth=1.75cm]{geometry}

\usepackage{amsmath}
\usepackage{graphicx}
\usepackage{longtable}
\usepackage[colorlinks=true, allcolors=blue]{hyperref}

\usepackage{todonotes}

\title{
Performance, Opaqueness, Consequences, and Assumptions: \\
Simple questions for responsible planning \\ 
of machine learning solutions}
\author{Przemyslaw Biecek \\ \small{MI$^2$.ai, Warsaw University of Technology}}

\begin{document}
\maketitle

\begin{abstract}
The data revolution has generated a huge demand for data-driven solutions. This demand propels a growing number of easy-to-use tools and training for aspiring data scientists that enable the rapid building of predictive models. 
Today, weapons of math destruction can be easily built and deployed without detailed planning and validation. This rapidly extends the list of AI failures, i.e. deployments that lead to financial losses or even violate democratic values such as equality, freedom and justice. 
The lack of planning, rules and standards around the model development leads to the ,,anarchisation of AI''. 
This problem is reported under different names such as validation debt, reproducibility crisis, and lack of explainability. Post-mortem analysis of AI failures often reveals mistakes made in the early phase of model development or data acquisition. Thus, instead of curing the consequences of deploying harmful models, we shall prevent them as early as possible by putting more attention to the initial planning stage. 

In this paper, we propose a quick and simple framework to support planning of AI solutions. The POCA framework is based on four pillars: Performance, Opaqueness, Consequences, and Assumptions. 
It helps to set the expectations and plan the constraints for the AI solution before any model is built and any data is collected. With the help of the POCA method, preliminary requirements can be defined for the model-building process, so that costly model misspecification errors can be identified as soon as possible or even avoided.
AI researchers, product owners and business analysts can use this framework in the initial stages of building AI solutions. 
\end{abstract}

\section{Introduction}

The data accessibility revolution has spawned a huge demand for data analysis. This demand is driving a growing number of tools,  courses, and activities to train new analysts and modellers. Hence the growing pressure for easy-to-use solutions that enable rapid model building. This development is taking place under the banner of ,,democratisation of AI'', which is supposed to reflect the inclusiveness of these tools. Anyone can be a model developer, just by pressing a few buttons or copying a few lines from a stack overflow.
But is this enough?
Does the lack of rules and restrictions around the model generation and validation really lead to the democratisation of ML methods?

Little is known about the standards of operation for modellers developing artificial intelligence solutions. Only fragmentary information based on interviews with analysts or online surveys studying large groups of users of systems such as Coursera or Kaggle can be found in the literature. However, the picture emerging from these interviews is quite gloomy. \cite{DataSharing2020} discusses low standards related to data sharing or keeping (mostly own laptop). Industry reports such as \cite{KaggleReport} focus on ML algorithms or the most commonly used modelling languages rather than a better understanding of the process of building safe and responsible ML solutions.

Perception of the model building process is often shaped by Kaggle-style competitions.
Kaggle\footnote{\url{https://www.kaggle.com/}} has been offering predictive modeling competitions since 2010. Over time, the name of the platform has become synonymous with prediction competitions in which numerous teams compete to build the ,,best'' predictive models, where,,best'' means with the highest performance on predefined validation data.
Under the cover of ,,fair'' and objective competition we get a well-defined problem and often well-defined objective function for optimisation. 
The team which obtains the highest performance on the hidden validation data wins.
Kaggle is a great platform for a wide community of data scientists with many learning materials and data sets available. However, there is also a dark side to its popularity, like the impression that the verification of the solution ends with a single validation set. When the results are announced, we do not find out if the model was deployed in the real world if its effectiveness was still high or degraded quickly, we do not know if the model behaved according to domain knowledge and if it was verified/debugged using XAI techniques. Solutions whose only verification is the calculation of a single criterion on validation data are accepted.  Since this is the most popular platform for novice data scientists, it may lead them to believe that model validation looks this way. Recently, the platform itself has been encouraging in-depth analysis of the problem, sharing materials related to the explainability, transparency and ethics of AI solutions \cite{KaggleE2021}. However, performance on the validation set remains king.

Among entry-level AI specialists, the idea of what AI solutions are is frequently shaped by events such as AI hackathons. AI hackathons are events inspired by software hackathons, i.e. usually 24 or 48-hour-long events during which teams compete to create the best solution to specific (sometimes not well-defined) problems.
Software hackathons are usually brainstorming types of events for quick prototyping of IT solutions. And they are great for prototyping solutions, networking, employee screening and so on.
Usually one of the criteria for evaluating the outcome is the presentation, during which the teams show the developed prototype demonstrating the potential capabilities of the target solution. 
In the case of AI solutions, value-added is often linked to the deployment of a developed model, so its effectiveness is difficult to assess well during a short hackathon.
This increases the temptation to use sales skills to tell what the model could do, although without new data and deep verification it is often difficult to know how realistic this is.
And with the magic of the term AI, it is easy to over-promise.
After analyzing several AI hackathons related to healthcare (see Appendix 1), one can see that in most cases the quality of a solution is evaluated solely based on one measure, often accuracy, rarely F1, very rarely AUC. Solutions are not subjected to expert evaluation and are not confronted with domain knowledge.
In some cases, the use of medical data is just a cover to spice up a competition that is organised for promotional or recruitment purposes.

According to \cite{KaggleReport2020} report, the main source of training materials for data scientists are courses from popular online platforms such as Coursera\footnote{\url{https://www.coursera.org/}}, edX\footnote{\url{https://www.edx.org/}} or Udemy\footnote{\url{https://www.udemy.com/}}.
And courses offered on these platforms are great. They provide huge opportunities for large numbers of people to learn new skills.
But when analysing the popularity of courses and specialisations related to deep learning or building AI systems, it is easy to notice that the vast majority of courses focus on the technical aspects of training models. There is just a small fraction of courses focused on in-depth analysis of models, transparency or bias.
MOOCs are certainly a huge and valuable source of material for developing new competencies. However, what is lacking in their current offerings is the ability to plan and critically evaluate the meaningfulness of building AI solutions. The situation is slowly changing, so one can find short courses on responsible innovation on large MOOC platforms \cite{EdxE2022}, or prepared by organisations working with AI models \cite{FastAi2020, LinuxE2021, AIE2022, EofAI2022}, but this is still only a small fraction of the materials on training new predictive models.

The aim of this work is not to criticise the status quo but to constructively propose a solution that can help to build better models. The review presented above shows that the perception associated with the AI analyst's work lacks a focus on the early planning stage.
To fill this gap in the next section, we present the POCA method, a set of simple questions that are worth tackling before you start building an AI solution. The answers to these four groups of questions will help to spot and eliminate common mistakes more quickly at the solution specification stage.

\clearpage

\section{POCA method: simple and practical questions on how to prepare for a development of AI solutions}

In this section, we present the POCA framework (an acronym for Performance, Opaqueness, Consequences, Assumptions) to increase the trustworthiness of predictive models via better planning.

The primary motivation behind this framework is the observation that today, in the modelling process, developers' almost entire focus is on performance measured on some test set. While such performance often does not translate into long-term performance or does not capture important issues related to model deployment such as bias of fairness.
To build responsible models, we need to expand our mindset and perspective to take into account other crucial threads and issues.

The POCA framework is designed to lead to more informed decisions regarding model development.
Like the SWOT analysis, the POCA framework also involves answering four key groups of questions. These groups are presented below, together with the supporting questions.

Questions presented below are selected to make the developer more aware of potential threats. But answers alone are not enough to reduce validation debt. Think about actions that shall be taken to optimize the future model behaviour.

\begin{figure}[h!]
    \centering
    \ 
    \\ 
    \
    \includegraphics[width=\textwidth]{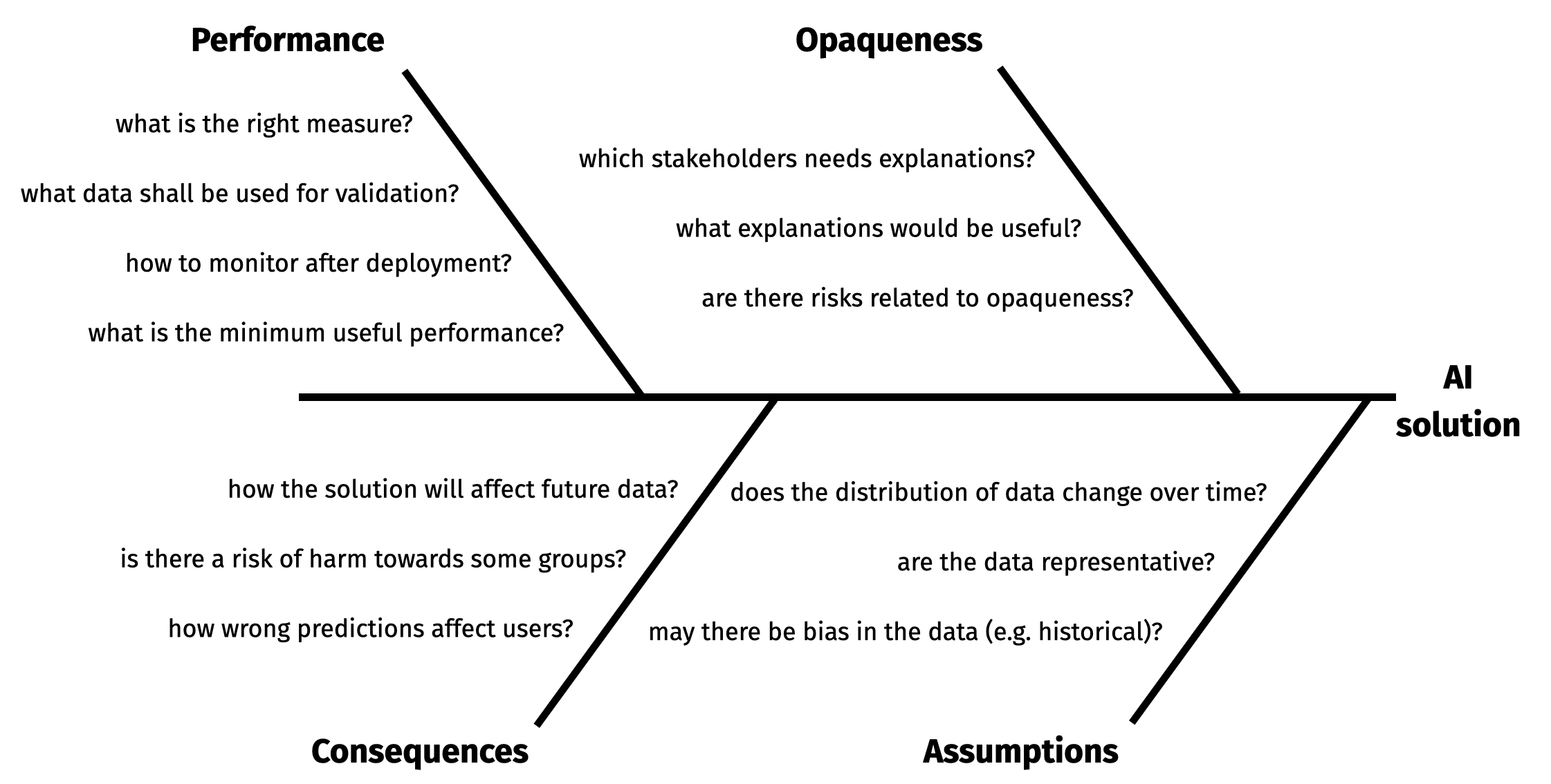}
    \caption{Ishikawa diagram for POCA framework for planning of trustworthy AI solutions. The supplementary questions help to look at each of the four criteria from multiple perspectives.}
    \label{fig:POCA}
\end{figure}

\subsection{Performance}

\begin{flushright}
\textit{Perhaps what you measure is what you get. 
\\
More likely, what you measure is all you’ll get. 
\\
What you don’t (or can’t) measure is lost.}
\\
H. Thomas Johnson
\end{flushright}

The management of anything, not just in data science, is strongly linked to quantitative progress indicators (KPIs). The experience of hundreds of thousands of managers shows that measuring is wildly important, but measuring the wrong things can be extremely harmful.

This also applies to the field of data science, where model training and model selection are often based on some ,,classical measures'' (MSE, accuracy, AUC) that may not be linked with business indicators.

This is why in the \textit{Performance} dimension we need to define how to measure success for the application of predictive models. Some specific questions that may help in this:

\begin{itemize}
    \item What is the right measure of model performance? Often models are optimized/validated using some default loss functions based on Accuracy or AUC/RMSE, even if they have nothing to do with the business objectives. During the planning phase, one should determine which metrics businesses care about. Usually, it is very hard because business applications are complex, but try to get as good measures as possible.
    \item What data shall be used for model validation?
    Clearly, we cannot use the same data for training and model validation. But quite often random split makes validation data too similar to training data and the use of random division into a learning set and a test set leads to overly optimistic conclusions (see \cite{Efron2020}).
    Think about different ways to generate validation data, like out-of-time validation (if there is a significant time component) or out-of-region (if there is a significant spatial component) or other validations that are not based on a purely random train-test split. 
    \item How to monitor the model after deployment? How frequently the performance shall be evaluated, and what criteria will alert us that the model is no longer working? Monitoring is critical to detecting data drift/model drift problems, or problem with model generalisation.
    \item What is the minimum useful performance? What is the minimum model quality that justifies its use? It is good to set the bar in advance because later it is very tempting to use solutions with close to the random performance. The more time we spend on modelling the more likely we are to use the model even if it is weak. Work against this risk and set the performance bare in advance.
    \end{itemize}

\subsection{Opaqueness}

\begin{flushright}
Linus’s Law aka \textit{Many Eyes Make All Bugs Shallow}
\\
Jeff Jones
\end{flushright}

After the rising adoption of Explainable Artificial Intelligence techniques, we are more prepared to analyse not only the model performance but also the model behaviour and sensitivity to data perturbations. There is also a long list of arguments why models are overly-opaque (see \cite{Rudin2019}). 
Not always full model transparency is possible and not everybody can understand the same explanations in the same way. But it is important to set our expectations related to the acceptable opaqueness of AI solutions.

Some specific questions that may help.

\begin{itemize}
    \item Which stakeholders need explanations? Probably not every stakeholder of every model needs to know how the model works, but certain groups of stakeholders may want/have the right/obligation to see how the model works. In the case of medical applications, perhaps the activity will be of interest to the physician, perhaps to the patient, or perhaps to the model auditor. Sometimes explainability to certain groups may be required by law or external regulations. Usually, the list of stakeholders is longer than just users and model developers. Think about other possible groups that may be interested or affected by the model. 
    \item What level of explanations is required by identified stakeholders? For each of these stakeholders, determine what kind of explanation would satisfy their curiosity. The stakeholders will have different knowledge of the domain, so they may expect different explanations. You should determine how detailed explanations will be and why.
    \item Are there risks related to opaqueness? If the model is complex and its full explanation is not possible, or the knowledge of a given stakeholder does not guarantee a full understanding of the model's operation. One should determine the risks stemming from this obfuscation. Is it better to use an original feature or some aggregates? Shall explanations be sparse or their accuracy is more important?
\end{itemize}

\subsection{Consequences}

\begin{flushright}
\textit{With great power comes great responsibility}
\end{flushright}

The main reason to develop predictive models is to improve or change some processes. Changes may have some obvious, not so obvious but expected and also some totally unexpected consequences. Spend some time thinking about the consequences of deployment of your model. How this will impact other processes or other stakeholders? Is there any direct or non-direct risk created by the model? If some consequences may be harmful to some stakeholders (lost jobs, lost medical privileges, higher costs) how then can be mitigated? It is very rare to introduce change in a way that everybody will benefit from it on the same scale. But think about consequences in short, medium and long time horizons. 

The following questions may help.

\begin{itemize}
    \item How the solution deployment will affect future data? Think about credit scoring, if a model gives credit only to those who will pay the credit, then we will have no new data about people in higher-risk groups. This may backfire in future.
    Is it possible, that a change in the data will lead to a need for model retraining? In what way the model deployment will affect the distribution of the data, and whether it is the desired way?
	\item Is there a risk of harm towards some groups? Can the deployment of a model lead to some protected group being disadvantaged? How you will monitor whether this is not happening?
	\item How incorrect predictions may affect different stakeholders? Most models give the wrong prediction from time to time, is there anything we can do to minimize the impact of false predictions, false positives/false negatives?
	\item What is desired opportunities? Think also about positive consequences to keep the eyes on the price. What positive change do you envision in with the AI solution? Who and how will benefit from it?
\end{itemize}

\subsection{Assumptions}

\begin{flushright}
\textit{Representative samples are all alike; every biased sample is biased in its own way.}
\\
Rephrased Leo Tolstoy
\end{flushright}

Training data is different from future data on which model will be deployed. Yet in most cases we do not have access to future data from future applications. So we need to assume that available data to some extent is similar to this future data. But what similarities between current training data and future data we can assume? 

The following questions may help to better understand the scope of similarity between past and future data.

\begin{itemize}
    \item Does the distribution of data change over time? Are sets of data that will feed the model in the future stationary or we can expect some periodicity (weekly, yearly) or some other kind of drift in data? What about other aspects like spatial or other factors?
    \item Is the sample representative? Is one training sample sufficient for all future purposes? The inability to construct a representative sample does not exclude the possibility of constructing a model, but it is certainly one of the questions that should be asked.
    \item May there be bias in the data (e.g. historical)? Can we do anything to minimize it?
\end{itemize}

Write down the answers to these questions. They will be helpful during the next stages of building and validating predictive models.

\clearpage

\section{Motivations that support the planning of AI systems}

The POCA framework is a synthesis of a set of other solutions to assist in strategic planning. Some of these solutions are almost directly transferable to the design of AI systems.

\subsection{SWOT analysis}

The SWOT analysis (see \cite{SWOTbook}) is a prevalent tool for strategic analysis of various undertakings. Like POCA analysis, SWOT analysis is also done at the planning stage, before the project is started. During this analysis, key questions are asked to determine: the key Strengths of the venture; Weaknesses of the venture; Opportunities that will open up; and potential Weaknesses. The SWOT framework is used for ventures of varying degrees of complexity, and it has the unquestionable advantage of looking at the venture more objectively from different key perspectives. The POCA framework has a similar objective. We determine if we are adequately prepared to build a responsible model by answering a series of key questions.

\subsection{Model Cards}

Model Cards (Model Cards for Model Reporting, see \cite{ModelCards2019}) is a tool for referring to the intended use of a model. It acts in a similar spirit to a model warranty by describing what applications the model can be used for. This is an important step towards responsible ML because previously pre-trained models were used without regard to the intentions and assumptions of the person training the model. The main difference between Model Cards and POCA is that Model Cards are a part of the documentation of the model. They are created once the model is trained and its effectiveness can be measured and the extent of usability determined. POCA is a framework to be used at the planning stage before we even start building the model and before we even collect data to develop the model. These tools complement each other, however, because elements of the POCA analysis can be used later when documenting the model using Model Cards

\subsection{Unknown unknowns}

The FairML book (see \cite{FairML}) has a significant influence on the POCA framework. It shows how thinking about fairness requires a broader perspective than just controlling for specific statistical coefficients. 
Problems identified in the post-mortem analysis of the model were unexpected.
The book makes it clear that the fairness of a model is often critically influenced by the quality of the data and its representativeness. At the same time, some bias may already be sewn into the data itself. Hence, responsible model thinking must look past the data, considering the sources of potential bias in the data.
But this is not enough because the model used in practice shapes some part of reality and can lead to bias in the future. Therefore, the POCA framework involves asking questions about both the past and the future and does not treat test data as the final frontier.

\subsection{Model Development Process}

 Model Development Process (see \cite{MDP2019})   is a framework that allows planning the resource allocation of individual tasks related to model building. The MDP framework helps plan the model's development to include all the necessary tasks, including the model building iterations and elements such as validation and testing. The difference between POCA and MDP is that with MDP we mainly think from the perspective of managing the resources needed to develop models, while with POCA we mainly plan constraints for the model taking into account perspectives of various stakeholders (usually related to  high-performance, transparency, fairness and safety of developed models)..

\section{Summary}

Practically every project can be divided into three main phases: planning, execution and retrospective of the results. Many educational resources related to data science focus on the ,,execution'' phase. They explain which tools/algorithms to use to build models. The increasing number of spectacular failures of AI solutions has increased the interest also in the retrospective stage of model analysis. Hence the growing interest in explainability and fairness in post-hoc analysis (in this paper referred to as post-mortem). To supplement these perspectives, this paper focuses on the first stage - planning. A well-conducted project should start with a detailed analysis of feasibility to avoid the most fundamental mistakes and misunderstandings.

The proposed framework deliberately consists only of four pillars, so that it can be carried out simply even for small projects. It is also not enough to answer these questions once at the beginning of the project, but it is worth to periodically monitor these answers during the project as we acquire new knowledge and experience.

Given these limitations, we hope that even such a simple framework will help to properly prepare for the development of AI solutions. And also will draw practitioners' attention to the initial stage of building AI solutions so that they are more tailored to the needs of future users.


\bibliographystyle{apalike}
\bibliography{sample}


\section{Annex 1}

\begin{longtable}{p{7cm}lp{8cm}l}
\hline
NAME                                                                              & YEAR & WEBSITE                                                                                               &  \\ \hline
Data 4 Healthy Recovery Hackathon                                                 & 2021 & https://eventornado.com/event/data-4-healthy-recovery\#home                                           &  \\
data for better health                                                            & 2019 & https://dataforbetterhealth.be/fair/                                                                  &  \\
MedHacks                                                                          & 2021 & https://www.medhacks.io/\#/                                                                           &  \\
Medical School Admissions                                                         & 2018 & https://www.kaggle.com/c/medical-school-admissions/overview                                           &  \\
Predicting Medical Costs                                                          & 2021 & https://www.kaggle.com/c/predicting-medical-costs/data                                                &  \\
TECH WEEKEND Data Science Hackathon                                               & 2019 & https://www.kaggle.com/c/tech-weekend-data-science-hackathon/overview/description                     &  \\
AAS-Hackathon                                                                     & 2019 & https://www.kaggle.com/c/aas-hackathon-may19/overview                                                 &  \\
bloods.ai Blood Spectroscopy Classification Challenge                             & 2021 & https://zindi.africa/competitions/bloodsai-blood-spectroscopy-classification-challenge                &  \\
To Vaccinate or Not to Vaccinate                                                  & 2021 & https://zindi.africa/competitions/zindiweekendz-learning-to-vaccinate-or-not-to-vaccinate             &  \\
Melanoma Tumor Size Prediction                                                    & 2020 & https://machinehack.com/hackathons/melanoma\_tumor\_size\_prediction\_weekend\_hackathon\_15/overview &  \\
Patient Drug-Switch Prediction                                                    & 2020 & https://machinehack.com/hackathons/patient\_drug\_switch\_prediction/overview                         &  \\
Flu Shot Learning: Predict H1N1 and Seasonal Flu Vaccines                         & 2021 & https://www.drivendata.org/competitions/66/flu-shot-learning/page/211/                                &  \\
DengAI: Predicting Disease Spread                                                 & 2021 & https://www.drivendata.org/competitions/44/dengai-predicting-disease-spread/                          &  \\
TissueNet: Detect Lesions in Cervical Biopsies                                    & 2021 & https://www.drivendata.org/competitions/67/competition-cervical-biopsy/                               &  \\
Alzheimer's Detection Challange                                                   & 2021 & https://www.aicrowd.com/challenges/addi-alzheimers-detection-challenge\#evaluation-criteria           &  \\
ImageCLEF 2018 Caption - Caption prediction                                       & 2018 & https://www.aicrowd.com/challenges/imageclef-2018-caption-caption-prediction                          &  \\
ImageCLEF 2019 Caption - Concept Detection                                        & 2019 & https://www.aicrowd.com/challenges/imageclef-2019-caption-concept-detection                           &  \\
ImageCLEF 2018 Tuberculosis - MDR detection                                       & 2018 & https://www.aicrowd.com/challenges/imageclef-2018-tuberculosis-mdr-detection                          &  \\
ImageCLEF 2018 Caption - Concept Detection                                        & 2018 & https://www.aicrowd.com/challenges/imageclef-2018-caption-concept-detection                           &  \\
ImageCLEF 2019 Tuberculosis - Severity scoring                                    & 2019 & https://www.aicrowd.com/challenges/imageclef-2019-tuberculosis-severity-scoring                       &  \\
ImageCLEF 2019 Tuberculosis - CT report                                           & 2019 & https://www.aicrowd.com/challenges/imageclef-2019-tuberculosis-ct-report                              &  \\
ImageCLEF 2020 Tuberculosis - CT report                                           & 2020 & https://www.aicrowd.com/challenges/imageclef-2020-tuberculosis-ct-report                              &  \\
MICCAI 2020: HECKTOR                                                              & 2020 & https://www.aicrowd.com/challenges/miccai-2020-hecktor\#evaluation-criteria                           &  \\
MICCAI 2021: HECKTOR                                                              & 2021 & https://www.aicrowd.com/challenges/miccai-2021-hecktor\#evaluation-criteria                           &  \\
AI Drug Discovery: Pharmacokinetic Parameter Prediction                           & 2019 & https://signate.jp/competitions/168\#evaluation                                                       &  \\
Chest XR COVID-19 detection                                                       & 2021 & https://cxr-covid19.grand-challenge.org/Rules/                                                        &  \\
MIDOG 2021 microscopy domain generalization challenge                             & 2021 & https://midog2021.grand-challenge.org                                                                 &  \\
Quantification of Uncertainties in Biomedical Image Quantification Challenge 2021 & 2021 & https://qubiq21.grand-challenge.org/QUBIQ2021/                                                        &  \\
Fetal Brain Tissue Annotation and Segmentation Challenge                          & 2021 & https://feta-2021.grand-challenge.org/Evaluation/                                                     &  \\
Breast Cancer Semantic Segmentation                                               & 2019 & https://bcsegmentation.grand-challenge.org                                                            &  \\
Brain Pre-surgical white matter Tractography Mapping challenge (BrainPTM) 2021    & 2021 & https://brainptm-2021.grand-challenge.org/BrainPTM-2021/                                              &  \\
The 2021 Kidney and Kidney Tumor Segmentation Challenge                           & 2021 & https://kits21.kits-challenge.org                                                                     &  \\
SIIM-ISIC Melanoma Classification                                                 & 2020 & https://www.kaggle.com/c/siim-isic-melanoma-classification/overview/evaluation                        &  \\
Covid-19 CT                                                                       & 2021 & https://covid-ct.grand-challenge.org/Evaluation/                                                      &  \\
A Knee MRI Dataset And Competition                                                & 2019 & https://stanfordmlgroup.github.io/competitions/mrnet/                                                 &  \\
APTOS 2019 Blindness Detection                                                    & 2019 & https://www.kaggle.com/c/aptos2019-blindness-detection/overview/evaluation                            &  \\
KNee OsteoArthritis Prediction (KNOAP2020) Challenge                              & 2020 & https://knoap2020.grand-challenge.org/Evaluation/                                                     &  \\
MICCAI 2019 Grand Pathology Challenge                                             & 2019 & https://digestpath2019.grand-challenge.org/Evaluation/                                                &  \\
PAIP 2019 Challenge                                                               & 2019 & https://paip2019.grand-challenge.org/Evaluation/                                                      &  \\
EAD Challenge: Multi-class artefact detection in video endoscopy                  & 2020 & https://ead2019.grand-challenge.org/Evaluation/                                                       & 
\end{longtable}

\end{document}